\documentclass[10pt, a4paper]{article}

\usepackage[]{lrec-coling2024} 

\usepackage{microtype}
\usepackage{graphicx} 
\usepackage{adjustbox} 
\usepackage{xspace}
\usepackage{amsfonts}
\usepackage{subfigure}
\usepackage{amsmath}
\usepackage{mathtools}
\usepackage{arydshln} 
\usepackage{multirow}
\usepackage{booktabs} 
\usepackage{makecell} 

\title{Word Order and World Knowledge}

\name{Qinghua Zhao$^1$, Vinit Ravishankar$^2$, Nicolas Garneau$^2$ and Anders Søgaard$^2$ } 

\address{$^1$Beihang University, $^2$University of Copenhagen \\
         zhaoqh@buaa.edu.cn,
         vinit.ravishankar@gmail.com\\
         nicolas.garneau@ift.ulaval.ca, 
         soegaard@di.ku.dk 
         }

\abstract{
Word order is an important concept in natural language, and in this work, we study how word order affects the induction of world knowledge from raw text using language models. We use word analogies to probe for such knowledge. Specifically, in addition to the natural word order, we first respectively extract texts of six fixed word orders from five languages and then pretrain the language models on these texts. Finally, we analyze the experimental results of the fixed word orders on word analogies and show that i) certain fixed word orders consistently outperform or underperform others, though the specifics vary across languages, and ii) the Wov2Lex hypothesis is not hold in pre-trained language models, and the natural word order typically yields mediocre results. 
The source code will be made publicly available at \url{https://github.com/lshowway/probing_by_analogy}.
 \\ \newline \Keywords{language model, fixed word order, world knowledge} }

\begin{document}

\maketitleabstract

\section{Introduction}
The distribution of dominant word orders is generally explained by communicative efficiency, e.g., dependency and information locality \cite{doi:10.1073/pnas.2122604119}, but what explains the fact that most languages exhibit some variation across different word orders? The standard theory is that different word orders coexist because of the influence of multiple languages and that there is pressure from language acquisition for fixing word order \cite{lupyan2022case}. Word order transfer between neighboring languages of course does not explain the current distribution of dominant word orders: the first language was probably structured using the Subject-Verb-Object syntax (SVO) \cite{doi:10.1073/pnas.1113716108}, but how did alternative word orders arise out of that? In this paper, we explore a novel hypothesis about the role of within-language word order variation: 

\begin{quote}
    {\bf The W{\sc ov}2Lex Hypothesis}\\Word order variation facilitates the acquisition of lexical semantics. 
\end{quote}

The W{\sc ov}2Lex hypohesis has considerable support from human language acquisition studies. Input variability is known to facilitate both first and second language acquisition \cite{Sinkeviciute2019}. \citet{55ab1728767a4cdea50b1464b37470f8}, for example, found that object variability facilitates new word learning. \citet{raviv2022} synthesized the above work, concluding that: 

\begin{quote}
{\small ``{An effective way of improving generalization is to expose learners to more variable (and thus often more representative) input. More variability tends to make initial learning more challenging but eventually leads to more general and robust performance. }''}
\end{quote}

We hereby present a series of experiments to check whether the W{\sc ov}2Lex hypothesis is still hold in pre-trained language models. Our experiments are inspired by \citet{sinha-etal-2021-masked} who pre-trained language models (LMs) on shuffled texts and compared their performance with models trained on the original one (natrual word order). \citet{sinha-etal-2021-masked} initially argued this did not lead to significant performance drops, but in subsequent work, \citet{abdou-etal-2022-word} identified limitations in \citet{sinha-etal-2021-masked}, and they were able to show that corpus perturbation leads to much lower performance. 

In this paper, we move a step from ``natural word order'' and ``shuffled word order'' toward ``fixed word order''. We first probe their LMs for world knowledge. We do so through word analogies, relying on the experimental protocols and datasets provided by \citet{garneau2021analogy}. Our results confirm previous findings: the performance of language models trained with shuffled word order exhibits only slight variations. We then pre-train our own LMs, following the protocol of \citet{sinha-etal-2021-masked} and \citet{abdou-etal-2022-word}, but on data with ``fixed'' word order rather than original (natural) or shuffled text.
We fix the word order by reordering the (subject, object, verb) items. We find that, surprisingly, fixing word order could lead to both performance increases or drops in different languages, while natural word order keeps more mediocre across all tested languages and relations.

\section{Related Work}
\paragraph{Why Word Order Variation?} Languages permitting word order variations typically use this variation to encode different pragmatic distinctions, such as referentiality, discourse anchoring, etc. \cite{Slioussar2011}. Others have argued that word order variation arises from optimization of processing (avoidance of ambiguity) and grammaticalization \cite{Levshina+2019+533+572}. Input variability is known to facilitate learning, including lexical knowledge \cite{55ab1728767a4cdea50b1464b37470f8,raviv2022}, but we are, to the best of our knowledge, the first to study the impact of {\em word order} variability on the induction of world knowledge. 

\paragraph{Word Order and Pre-training} \citet{sinha-etal-2021-masked,hessel-schofield-2021-effective} pre-trained several LMs from scratch with shuffled text, relying on various shuffling strategies, such as bi-gram order permutation. They concluded from their experiments that word order information matters little for the downstream performance~\cite{wang-etal-2018-glue}.
However, \citet{abdou-etal-2022-word} tried to replicate their experiment and found a limitation in their work.
The shuffling strategy they used did not affect the positional encodings of input tokens, and when shuffling at the sub-word level, the  downstream performance dropped significantly. 

\paragraph{World Knowledge and Word Analogies} World knowledge~\cite{clark-etal-2007-role} refers to the information about the real world that individuals accumulate over time. It encompasses facts and concepts about the broader world. Word analogy~\cite{ulcar-etal-2020-multilingual} is a comparison between two things, typically on the basis of their relations. For example, ``Beijing is to China as Washington is to the USA''. Word analogies can test our understanding of the world. Since language models are trained on a vast amount of text, we adopt a word analogy dataset that is extracted from Wikidata to test the relations between word order and world knowledge. 

\section{Experiments}
To investigate the effects of word order on world knowledge, we conduct tests on language models that are trained on corpora with different word orders, including the original (natural) order, shuffled order, and fixed word order.

\paragraph{Natural Word Order.}
For models trained on corpora of natural texts, we use the original BERT$_{\text{base-uncased}}$, mBERT$_{\text{base-uncased}}$ \cite{devlin-etal-2019-bert} and RoBERTa$_{\text{base}}$ \cite{liu2019roberta} LMs. They have 12 layers with 12 attention heads, and RoBERTa is an extended version of BERT with more pre-training corpus and training steps, i.e., pre-trained on a corpus of 160GB for 500K steps. We refer to these models trained on natural word order as ``natural models''.

\paragraph{Shuffled Word Order.}
To assess the significance of word order in pre-training, \citet{sinha-etal-2021-masked, abdou-etal-2022-word} pre-trained RoBERTa models using corpora that have been shuffled at various levels, such as corpus-level and unigram- or bigram-level, or at different phases such as before or after tokenization. 
These models are referred to as ``shuffled models''. In this paper, we also examine their word analogies.
Specifically, we test models \textit{\texttt{shuf.n1}}, \textit{\texttt{shuf.n2}}, \textit{\texttt{shuf.n3}}, \textit{\texttt{shuf.n4}}, \textit{\texttt{shuf.cps}} and \textit{\texttt{nopos}}  released by \citet{sinha-etal-2021-masked}, which are pre-trained from scratch using the Toronto Book Corpus and English Wikipedia (16GB) on a full-scale RoBERTa base model. The models were trained for 100,000 steps over 72 hours using 64 GPUs.
\textit{\texttt{shuf.n1}} $\sim$ \textit{\texttt{shuf.n4}} refer to LMs  pre-trained  on the shuffled text at $n$-gram level (please refer to \citet{sinha-etal-2021-masked} for details). The \textit{\texttt{shuf.cps}} refers to LMs pre-trained on an entire reshuffled corpus (i.e., each word is sampled according to its frequency). The \textit{\texttt{nopos}} refers to the LMs being pre-trained from scratch without positional embeddings, since positional embeddings encode the sequence order information. 

\paragraph{Fixed Word Order.} 
We further pre-train RoBERTa$_{\text{base}}$ models using corpora adhering to fixed word orders. While there are multiple approaches to define fixed word orders—ranging from alphabetical to word frequency-based, or even based on parts of speech—our primary objective is to  investigate whether natural word order yields superior representations in language models compared to fixed word order. Consequently, our primary interest lies in examining word orders that occur naturally in languages, such as SVO (Subject-Verb-Object), OVS (Object-Verb-Subject), and the like. 

Specifically, we first extract the dependency tree of Wikipedia text using SpaCy, where named entities and noun chunks are merged (if it was implemented in SpaCy). Instead of traversing the dependency parse tree to avoid complex regular expressions, we use a data-driven method motivated by Word2Vec \cite{mikolov2013efficient} to get $n$-grams.
We then check each $n$-gram and extract the subject, object, and verb item, rearrange the positions of the subject, object, and verb within the extracted $n$-grams to obtain $n$-grams of different word orders, including SVO, SOV, VOS, VSO, OSV, and OVS. 
Finally, these reordered $n$-grams are concatenated to create pre-training corpora, which is identical except for the word order.
The extracted $n$-grams, without any ordering modifications, are referred to as the natural word order corpus, i.e., \textit{\texttt{fixed.ntr}} in Table \ref{fixed_results}.

To obtain enough training text segments, we set $n$=5 for English, German, French, Spanish, and $n$=10 for Polish as the merge of named entities and noun chunks is not implemented for Polish in SpaCy. If there are multiple subjects/objects/verbs in the $n$-gram, only the first one is used to avoid duplicate. In total, we obtain pre-training corpora for English, German, French, Spanish and Polish, consisting of 2 million, 2 million, 0.5 million, 1 million and 1.5 million $n$-gram, respectively. Note that, SVO refers to a kind of word order, and \textit{SVO} refers to the corresponding trained LM.

To pre-train the language models, we set the training batch size to 16, and the accumulation step to 8. The learning rate is set to 1e-4, and we use the AdamW optimizer with a linear learning rate scheduler, with warmup step of 1\% of the total training step.
We use 4 GPUs, and train them for 50,000 to 10,000 steps with early stopping. All other parameters are kept as their default values.

\paragraph{Random initialization. } 
As a lower bound, we also test the performance of randomly initialized tokens' embeddings.
Specifically, for RoBERTa, instead of using the pre-trained embedding table, we initialize the embedding table with  $\text{mean}=0$, $\text{std}=0.02$, denoted as \textit{\texttt{R-rand}} as shown in Figure \ref{natural_shuffled_models}. We also test the performance of Fasttext \cite{bojanowski-etal-2017-enriching}, as shown in Table \ref{overall_results}.

\paragraph{Tasks and Datasets.} To study the impact of word orders on word analoies, we use the WiQueen dataset, as introduced by \citet{garneau2021analogy}. WiQueen is a publicly available, Wikidata-based, large-scale, multilingual (11 languages) multi-word analogy dataset. This dataset is an analogy retrieval task, which is a type of natural language processing task that aims to find an analogous relationship between a source concept and a target concept.
We analyze the combined training, evaluation, and test datasets of WiQueen, yielding a total of 4,815 samples across 139 relationships and 10,124 concepts. Following the original setting, we evaluate the performance by Precision@1.

\subsection{Analysis of Natural/Shuffled Word Order}\label{exp:natural_fixed}
\begin{figure}[!ht]
\centering
\includegraphics[width=\columnwidth]{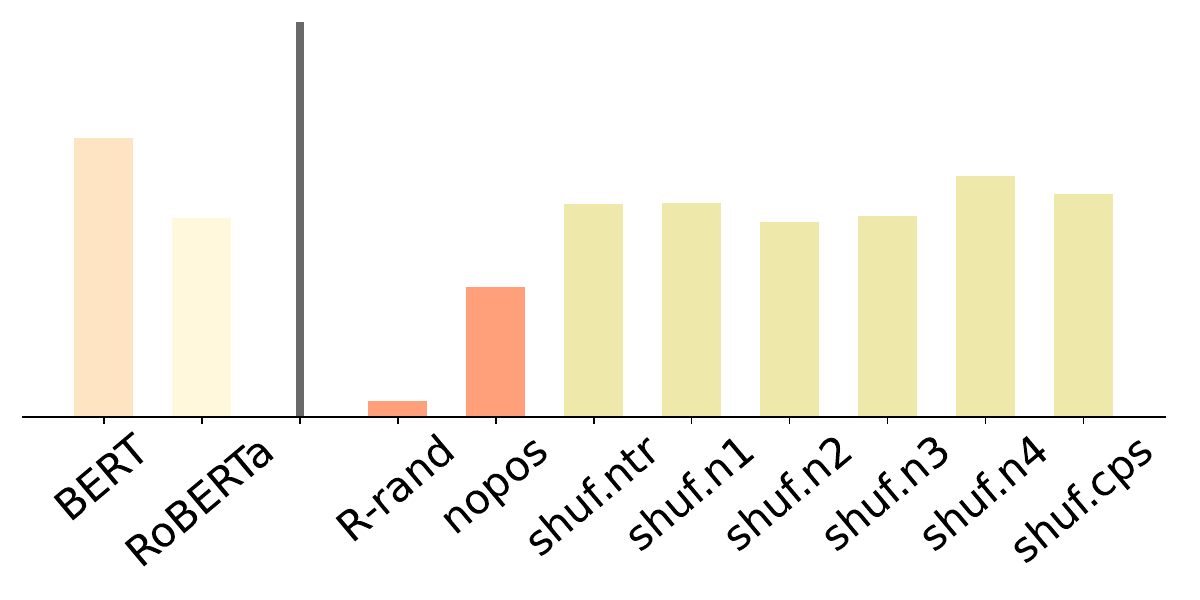}
\caption{The results of natural and shuffled models on WiQueen. Table \ref{natural_shuffled_models} shows the specific numbers.}
\label{natural_shuffled_models}
\end{figure}

Since the relation between word order and word analogy has not been explored, here, we re-test the performance of existing natural and shuffled LMs by conducting experiments on WiQueen dataset. 
As shown in Figure \ref{natural_shuffled_models}, compared with the natural word order \textit{\texttt{shuf.ntr}}, \textit{\texttt{nopos}} and \textit{\texttt{R-rand}} result in a substantial decrease in performance, it is easy to understand. However, when comparing \textit{\texttt{shuf.ntr}} with \textit{\texttt{shuf.n1}} $\sim$ \textit{\texttt{n4}} and \textit{\texttt{shuf.cps}}, same as existing results, we also find that the permutation of word order lead to either a decrease or an increase in performance with slight differences. Several hypotheses attempt to elucidate the results. For instance, it might be that the tasks tested do not require word order information, or that the evaluated language models do not rely heavily on word order information. Another one suggests that the shuffling methods employed might not destroy essential word order information. However, these hypotheses have yet to be widely validated. 

\subsection{Analysis of Fixed Word Order}
\begin{figure}[!t]
\centering
\includegraphics[width=\columnwidth]{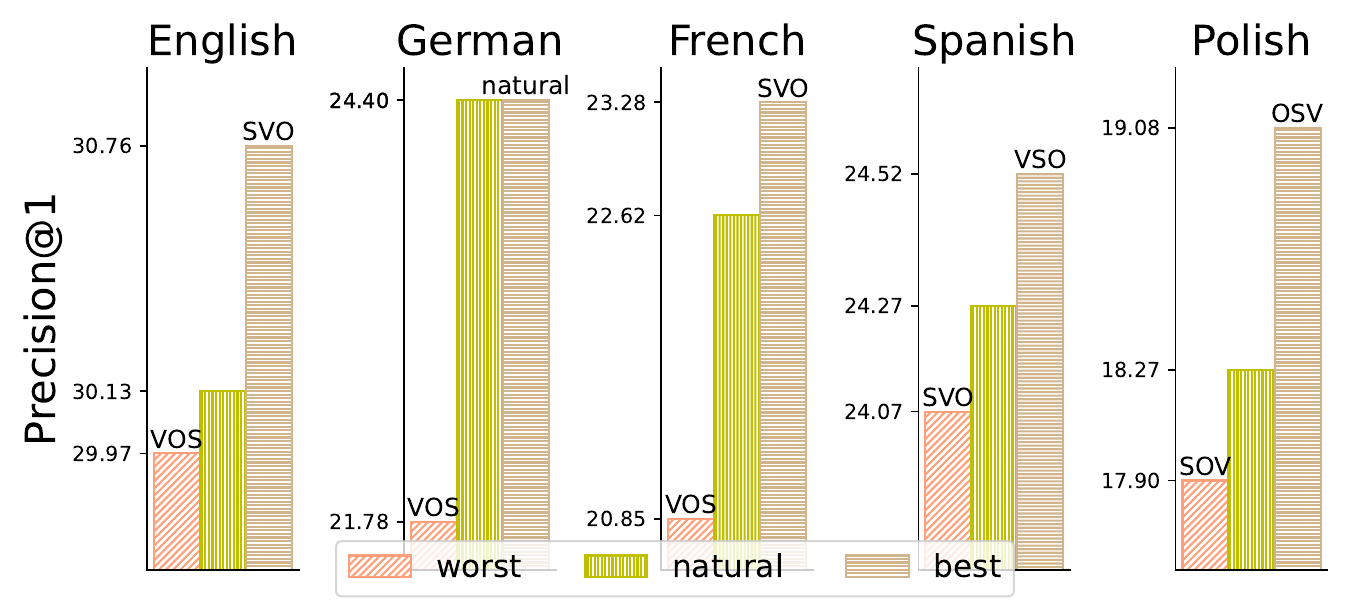}
\caption{Results of worst/best fixed word orders and natural word order. Table \ref{fixed_results} shows the specific numbers.}
\label{orders_steps_en_fr}
\end{figure}

Building upon the work in \citet{sinha-etal-2021-masked, abdou-etal-2022-word}, we extend the exploration of word order from ``natural'' and ``shuffled'' to ``fixed'' word order.  We pre-train RoBERTa$_{\text{base}}$ from scratch using pre-training corpora that have been reordered to different word orders. Across these corpora, all input items are the same except the positions of the subject, verb, and object.  Figure \ref{orders_steps_en_fr} illustrates the Precision@1 results, with these experiments, we try to answer the following questions: 
\begin{enumerate}
    \item Q1: Does a particular word order exhibit superiority or inferiority  to the others?
    \item Q2: Does the use of hybrid word orders (i.e., natural word order) confer any advantages over the use of a single word order?
    \item Q3: Does the question Q1 hold consistent across different languages?
\end{enumerate}

In answer to Q1, the results are positive. There is a clear distinction between the best and worst fixed word orders across all examined languages. Besides, the \textit{SVO} performs the best in English and French but worst in Spanish. On the other hand, \textit{VOS} performs the worst in English, German and French. For the tested five languages, the first three (English, German, French) demonstrates similar trends, and show completely different trends with Spanish or Polish.

In response to question Q2, the findings are in the negative. Our analysis indicates that integrating a diverse range of word orders (i.e., natural word order), doesn't offer marked enhancements over individual, fixed word orders. However, it's important to highlight that, relative to other word sequences, the natural word order consistently exhibits more resilience across languages, maintaining a middle-ground performance.

In relation to question Q3, the answer leans toward the negative. The data suggests the presence of both superior and inferior word orders among the seven considered (six pre-defined and one natural). However, this hierarchy varies across languages. Specifically, German and French demonstrate tendencies more aligned with English than do Polish and Spanish. Consequently, these findings do not entirely corroborate the {\sc Wov}2Lex hypothesis.

These empirical findings underscore the significance of word order, challenging prior results detailed in Section~\ref{exp:natural_fixed} (i.e., the permutation of word order lead a marginal decrease or increase in performance). While currently it's challenging to provide definitive explanations for these observations, we hypothesize that language models and linguistics may possess different ways to comprehend and process natural language.

\section{Other Findings}\label{other_findings}
We also have additional empirical findings that are not related to word order. Referring to Figures \ref{natural_shuffled_models} and \ref{orders_steps_en_fr}, with detailed data provided in Tables \ref{overall_results} and \ref{fixed_results}. The performance may be expected  to be RoBERTa $>$ BERT $>$ \textit{\texttt{fixed.ntr}}, where $>$ signifies superior performance.
Because RoBERTa and BERT have the same model architecture\footnote{Although BERT also uses next sentence prediction as a pre-training task, it is claimed that it matters little.}, and compared with the original BERT, RoBERTa is trained with more training corpus and longer training time, and \textit{\texttt{fixed.ntr}} is trained with a smaller corpus and fewer steps.
Contrary to expectations, the results reveal a sequence of \textit{\texttt{fixed.ntr}} $>$ BERT $>$ RoBERTa. Illustrating with English data, the performance metrics are $30.13 > 28.0 > 21.16$. 
This discrepancy might arise because the Wikipedia corpus alone suffices for the WiQueen dataset, given its derivation from Wikidata. Introducing additional corpora, like the Toronto Book Corpus, could potentially be detrimental to the data, leading to issues such as catastrophic forgetting.

\subsection{Word Order on Different Relations}
\begin{figure}[t]
\centering
\includegraphics[width=\columnwidth]{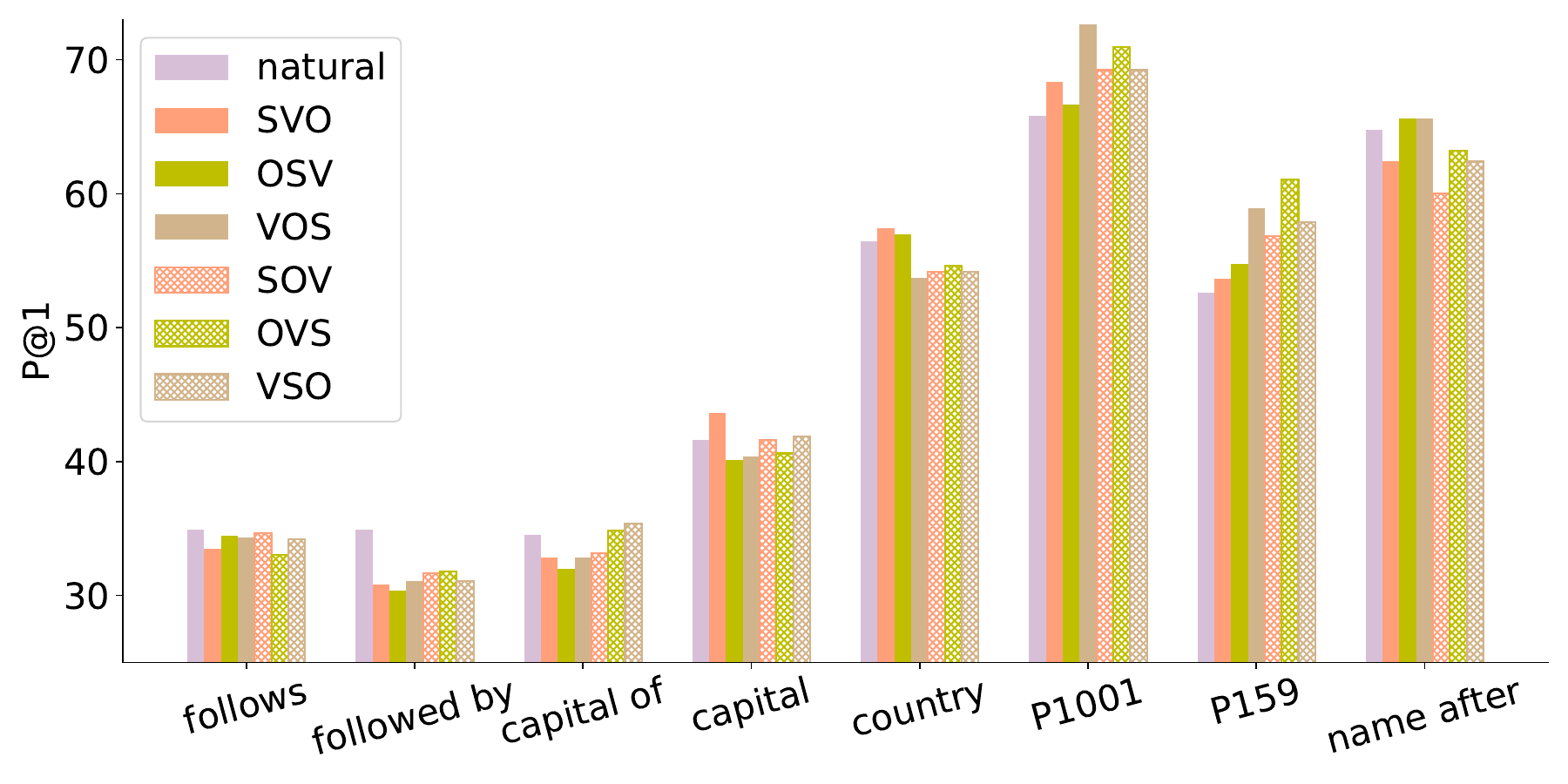}
\caption{Performance on different relations. ``P1001'' refers to relation ``applies to jurisdiction'', and ``P159'' refers to relation ``headquarters location''. Table \ref{fixed_relations_results} shows the specific numbers.}
\label{different_relations}
\end{figure}

In order to examine the impacts of ``fixed'' word order models on analogy relations, such as the relation ``capital'' in the example ``Beijing is to China as Washington is to the United States'', we evaluate the eight most frequently occurring relations. 
As shown in Figure~\ref{different_relations}, the natural word order demonstrates superior performance on certain relations such as ``follows'', ``followed by'', and ``capital of'' even though language models have lower scores on these relations. However, it performs poorly on the relations ``P1001'' and ``P159''. Meanwhile, \textit{SVO} is better at relations such as ``capital'' and ``country'', and the \textit{VOS} has slightly better performance on the relation ``P1001''.
On the other hand, models that start with \textit{S} such as \textit{SOV} and \textit{SVO}, exhibit the poorest performance on the ``name after'' relation.
The results indicates that models trained on different single word orders (SVO, SOV, VOS, VSO, OVS, OSV) and mixture word order (natural) exhibit varying abilities on different relations.

\section{Conclusion}
The objective of this paper is to investigate the influence of word order on world knowledge, including verifying whether {\sc Wov}2Lex hypothesis is supported in pre-trained language models. To accomplish this, we pre-trained language models from scratch on corpora with six fixed word orders, and then assessed their performance on an analogy dataset. 
The findings reveal that  natural word order does not consistently outperform the six tested fixed word orders. This suggests that the Wov2Lex hypothesis is not entirely supported by our empirical data. We hypothesize that this disparity arises from the distinct processing methods used in linguistics versus language models.

\bibliographystyle{lrec-coling2024-natbib}
\bibliography{custom,anthology}

\appendix\label{sec:appendix}
\section{Appendix}

\subsection*{Limitations}
The definition of fixed word order encompasses a broad range of possibilities, including alphabetical order, orders based on frequencies, nouns, verbs, adverbs, quantifiers, etc., and even subword-level order. In our exploration, we focus on a specific aspect of fixed word order, selecting 5/10-grams that contain a subject, object, and verb. Besides, we do not delve into other sizes of n-grams, such as 3-grams or 30-grams, due to considerations regarding carbon emissions.
The experimented languages are from similar language families and typologies.

\begin{table*}[h]
\centering
\resizebox{\textwidth}{!}{
\begin{tabular}{llll|ll|lllllllllll}
\multirow{2}{*}{\textbf{Language}}&  & \multicolumn{6}{c}{\textbf{Original}}&  & \multicolumn{7}{c}{\textbf{Shuffled}}    \\
\cline{3-8}
\cline{10-17}
 & & \textit{F-rand}& \textit{Fasttext}&   \textit{BERT}&   \textit{mBERT}&  \textit{RoBERTa}&  \textit{R-rand}&  &   \textit{shuf.ntr}&    \textit{ shuf.n1}&   \textit{shuf.n2}&   \textit{shuf.n3}&   \textit{shuf.n4}&   \textit{shuf.cps}&    & \textit{nopos}   \\
\cline{3-8}
\cline{10-15}
\cline{17-17}
Danish&     &    0.0&    13.02&    16.24&    17.51&    12.19&    2.64&     &    14.21&    13.29&    12.02&    11.11&    13.77&    15.08&     &    6.9	\\
German&     &    0.0&    10.67&    17.36&    22.7&    13.44&    3.7&     &    13.58&    13.83&    13.21&    13.94&    15.78&    14.58&     &    11.26	\\
English&     &    0.0&    10.96&    28.0&    29.7&    21.16&    1.7&     &    19.31&    20.62&    20.21&    20.64&    23.99&    16.57&     &    15.12	\\
Spanish&     &    0.0&    8.65&    19.21&    24.13&    13.31&    1.29&     &    14.5&    14.91&    13.21&    14.91&    16.8&    15.97&     &    9.93	\\
Finnish&     &    0.0&    12.59&    14.91&    16.88&    11.44&    2.55&     &    13.54&    12.77&    12.0&    11.44&    12.42&    12.71&     &    7.56	\\
French&     &    0.0&    6.72&    18.44&    21.18&    12.38&    4.92&     &    13.42&    13.29&    13.23&    13.29&    15.26&    15.97&     &    10.09	\\
Italian&     &    0.0&    7.13&    17.15&    22.58&    12.46&    4.28&     &    12.96&    13.54&    11.84&    13.06&    15.18&    14.43&     &    9.03	\\
Dutch&     &    0.0&    10.56&    17.01&    21.68&    12.88&    1.77&     &    14.5&    14.75&    12.92&    12.69&    16.49&    15.85&     &    9.87	\\
Polish&     &    0.0&    11.07&    13.54&    17.07&    10.97&    4.71&     &    11.53&    11.51&    10.28&    10.43&    12.81&    11.78&     &    6.96	\\
Portuguese&     &    0.0&    8.28&    15.99&    17.69&    12.07&    1.14&     &    12.42&    12.25&    10.43&    11.69&    13.37&    12.88&     &    7.06	\\
Swedish&     &    0.0&    9.80&    15.51&    17.67&    11.96&    3.24&     &    12.81&    12.4&    12.15&    11.84&    13.89&    12.69&     &    7.71	\\
\cline{1-17}
\textbf{Averages}&    &    0.0&    9.95&    17.58&    20.8&    13.11&    2.9&    &    13.89&    13.92&    12.86&    13.19&    15.43&    14.41&    &    9.23	\\
\bottomrule
\end{tabular}
}
\caption{Evaluation of natural, shuffled word order on the WiQueen dataset. \textit{F-rand} refers to \textit{Fasttext} with random initialization.} 
\label{overall_results} 
\end{table*}

\begin{table*}[h]
\centering
\resizebox{\textwidth}{!}{
\begin{tabular}{llll|lllllllllcll}
\multirow{2}{*}{\textbf{Relation}}&  & \multicolumn{4}{c}{\textbf{Original}}&  & \multicolumn{8}{c}{\textbf{Shuffled}}    \\
\cline{3-6}
\cline{8-16}
 & &    \textit{BERT}&   \textit{mBERT}&  \textit{RoBERTa}&  \textit{R-rand}&  &  \textit{shuf.ntr}&    \textit{shuf.n1}&  \textit{ shuf.n2}&   \textit{shuf.n3}&   \textit{shuf.n4}&   \textit{shuf.cps}&  $\Delta$ & & \textit{nopos}   \\
\cline{3-6}
\cline{8-14}
\cline{16-16}
followed by&     &    34.6&    32.91&    27.68&    3.11&     &    31.64&    27.4&    27.4&    26.27&    31.5&    29.24&    3.28&     &    24.01	\\
follows&     &    35.8&    34.49&    26.81&    10.58&     &    27.39&    26.52&    26.67&    26.52&    31.01&    29.71&    1.68&     &    22.32	\\
P190&     &    2.88&    2.88&    2.52&    0.36&     &    3.06&    2.52&    3.24&    1.98&    2.88&    2.52&    0.5&     &    2.7	\\
capital&     &    35.53&    38.58&    25.89&    2.79&     &    23.1&    29.7&    26.65&    27.92&    26.9&    22.34&    3.91&     &    21.83	\\
capital of&     &    30.62&    32.3&    22.75&    5.06&     &    18.82&    25.56&    21.35&    25.56&    27.53&    20.22&    5.22&     &    17.98	\\
P131&     &    34.75&    39.41&    34.75&    1.27&     &    24.58&    30.51&    32.2&    30.51&    33.05&    22.88&    5.93&     &    24.15	\\
country&     &    51.85&    55.09&    33.33&    2.31&     &    29.63&    31.94&    35.65&    36.57&    37.04&    16.2&    7.22&     &    16.2	\\
name after&     &    46.4&    52.0&    37.6&    0.0&     &    23.2&    30.4&    36.0&    36.0&    39.2&    25.6&    10.24&     &    28.0	\\
P1001&     &    57.26&    58.97&    28.21&    4.27&     &    17.09&    30.77&    27.35&    28.21&    37.61&    9.4&    12.65&     &    17.09	\\
headquarters location&     &    45.26&    54.74&    35.79&    4.21&     &    32.63&    45.26&    37.89&    40.0&    43.16&    25.26&    8.63&     &    33.68	\\
\cline{1-16}
\textbf{Averages}&    &    37.49&    40.14&    27.53&    3.4&    &    23.11&    28.06&    27.44&    27.95&    30.99&    20.34&    5.93&    &    20.8	\\

\bottomrule
\end{tabular}
}
\caption{Evaluation on different relations on WiQueen dataset. ``P190'' is the relation ``twinned administrative body'',  ``P131'' is the relation ``located in the administrative territorial entity'',  ``P1001'' is the relation ``applies to jurisdiction''. $\Delta$ refers to the average differences between \textit{\texttt{shuf.n1}} $\sim$ \textit{\texttt{shuf.n4}}, \textit{\texttt{shuf.cps}} and \textit{\texttt{shuf.ntr}}.} 
\label{relation_results} 
\end{table*}

\begin{table}[h]
\centering
\resizebox{\columnwidth}{!}{
\begin{tabular}{lllllllll}
\multirow{2}{*}{\textbf{Language}}&   &    \multicolumn{7}{c}{\textbf{Fixed}}    \\
\cline{3-9}
 & &  \textit{SVO}&  \textit{SOV}& \textit{OSV}&   \textit{OVS}& \textit{VSO}&    \textit{VOS}&     \textit{fixed.ntr}  \\
 \cline{3-9}
English&    &   30.76&  30.7&   30.09&  29.99&  30.3&   29.97&  30.13\\
German& &   23.9&  24.15&  22.9&   22.96&  24.38&  21.78&  24.4\\
French& &   23.28&  22.82&  23.18&  22.85&  22.93&  20.85&  22.62\\
Polish& &   18.54&  17.90&  19.08&  18.13&  18.25&  18.27&  18.27 \\
Spanish&    &   24.07&  24.19&  24.27&  24.27&  24.52&  24.09&  24.27\\

\bottomrule
\end{tabular}
}
\caption{Evaluation of fixed word orders on the WiQueen dataset. The best performance is reported.} 
\label{fixed_results} 
\end{table}

\begin{table}[h]
\centering
\resizebox{\columnwidth}{!}{
\begin{tabular}{lllllllll}
\multirow{2}{*}{\textbf{Relation}}&   &    \multicolumn{7}{c}{\textbf{Fixed}}    \\
\cline{3-9}
 & &  \textit{SVO}&  \textit{VOS}& \textit{OSV}&   \textit{SOV}& \textit{VSO}&    \textit{OVS}&     \textit{fixed.ntr}   \\
 \cline{3-9}
followed by&     &    30.79&    31.07&    30.37&    31.64&    31.07&    31.78&    31.64	\\
follows&     &    33.48&    34.35&    34.49&    34.64&    34.2&    33.04&    34.93	\\
P190&     &    3.96&    2.52&    3.06&    2.7&    3.42&    3.24&    3.24	\\
capital&     &    43.65&    40.36&    40.1&    41.62&    41.88&    40.61&    41.62	\\
capital of&     &    32.87&    32.87&    32.02&    33.15&    35.39&    34.83&    34.55	\\
P131&     &    39.41&    40.68&    40.68&    38.14&    39.83&    38.98&    40.68	\\
country&     &    57.41&    53.7&    56.94&    54.17&    54.17&    54.63&    56.48	\\
name after&     &    62.4&    65.6&    65.6&    60.0&    62.4&    63.2&    64.8	\\
P1001&     &    68.38&    72.65&    66.67&    69.23&    69.23&    70.94&    65.81	\\
headquarters location&     &    53.68&    58.95&    54.74&    56.84&    57.89&    61.05&    52.63	\\

\bottomrule
\end{tabular}
}
\caption{Evaluation of fixed word order models on different relations. } 
\label{fixed_relations_results} 
\end{table}

\end{document}